\documentclass[journal]{IEEEtran}
\usepackage{graphicx}
\usepackage{amsmath}
\usepackage{amssymb}
\usepackage{booktabs}
\ifCLASSINFOpdf
\else
\fi
\begin{document}
%
\title{Node Classification via Semantic-Structural Attention-Enhanced Graph Convolutional Networks}
%
%
%

\author{Hongyin~Zhu
\thanks{e-mail: hongyin\_zhu@163.com}
}

%
%

\markboth{Journal of IEEE}%
{Shell \MakeLowercase{\textit{et al.}}: Bare Demo of IEEEtran.cls for IEEE Journals}
%



\maketitle

\begin{abstract}
Graph data, also known as complex network data, is omnipresent across various domains and applications. Prior graph neural network models primarily focused on extracting task-specific structural features through supervised learning objectives, but they fell short in capturing the inherent semantic and structural features of the entire graph. In this paper, we introduce the semantic-structural attention-enhanced graph convolutional network (SSA-GCN), which not only models the graph structure but also extracts generalized unsupervised features to enhance vertex classification performance. The SSA-GCN's key contributions lie in three aspects: firstly, it derives semantic information through unsupervised feature extraction from a knowledge graph perspective; secondly, it obtains structural information through unsupervised feature extraction from a complex network perspective; and finally, it integrates these features through a cross-attention mechanism. By leveraging these features, we augment the graph convolutional network, thereby enhancing the model's generalization capabilities. Our experiments on the Cora and CiteSeer datasets demonstrate the performance improvements achieved by our proposed method. Furthermore, our approach also exhibits excellent accuracy under privacy settings, making it a robust and effective solution for graph data analysis.
\end{abstract}


%
\IEEEpeerreviewmaketitle

\section{Introduction}
Graph theory finds diverse applications across various domains, encompassing social networks, the Internet of Things, integrated circuits, recommendation systems, knowledge graphs, biomolecular networks, communication networks, and academic citation networks, all of which constitute graph data. The task of node classification involves classifying nodes based on their characteristics and the underlying network structure within graph data. This classification is crucial for various applications, such as identifying key information, enhancing computing performance, fault diagnosis, and analyzing network characteristics. The majority of existing machine learning models are primarily designed to handle Euclidean structured data, encompassing images, text, and array data. However, graph data exhibits a non-Euclidean structure, posing significant challenges for traditional machine learning methods to model effectively. Graph neural networks emerge as a promising approach, leveraging deep learning techniques to process graph data, automatically extracting features, and performing graph classification. 

Previous research has advocated the utilization of graph convolutional networks (GCNs) \cite{DBLP:conf/iclr/KipfW17} and graph attention networks (GATs) \cite{velickovic2017graph} for node classification tasks. These methods typically depend on task-specific graph data features, which are learned by neural networks and extracted through loss functions and supervised learning techniques. However, a notable limitation of this approach is that it focuses solely on a single type of feature, overlooking other crucial structural and semantic features within graph data that cannot be captured by neural networks alone. These overlooked features play a pivotal role in enhancing the generalization performance of the model. In this paper, we aim to address this limitation by harnessing the rich structural and semantic features inherent in the entire graph data, enabling the graph convolutional network to leverage them effectively. We hope to bolster the generalization capabilities of the graph convolutional network and enhance its accuracy in classifying nodes across diverse graphs. 

Leveraging inherent features within graph data is a fundamental aspect of graph neural network development. While graph convolutional networks typically rely on supervised learning tasks to extract features, these extracted features are often closely tied to the specific architecture of the model. Conversely, graph embedding algorithms strive to learn a representation of nodes in graph data that is agnostic to downstream tasks. Through unsupervised means, these algorithms aim to map nodes with similar structural properties to nearby points in a vector space. In this study, we employ the node2vec algorithm \cite{grover2016node2vec} to delve into the underlying structural characteristics of graph data. Specifically, we harness the power of biased random walks to learn informative node embeddings. We hope to capture the rich structural information encoded in the graph data, enabling more accurate and robust node classification.

Knowledge graph embeddings have demonstrated remarkable performance on tasks such as link prediction in graph data. Apart from graph structural features, the semantic relations between nodes are also paramount in node classification tasks. To tackle this challenge, we introduce a novel framework, semantic-structural attention-enhanced graph convolutional networks (SSA-GCN). This approach innovatively integrates semantic features with the structure of graph data through a cross-attention mechanism. By effectively combining the strengths of both semantic and structural features, SSA-GCN strives to enhance the overall performance in graph data processing. Furthermore, our proposed model has the potential to significantly improve the parameter optimization process of graph models.

The SSA-GCN leverages unsupervised learning algorithms to enrich the input-stage node features, enabling the model to acquire valuable external knowledge that is pivotal for node classification tasks. By seamlessly integrating the implicit representation of this knowledge into the model, SSA-GCN captures information that the model cannot independently learn, thereby enhancing its parameter optimization process. Specifically, we employ random walk and Skip-gram algorithms to capture the structural knowledge of nodes, while knowledge graph embedding provides semantic knowledge to the model. Furthermore, our method supports node classification tasks in privacy-preserving settings, even when the features of nodes are unavailable, demonstrating its versatility and adaptability.

In this paper, we present comprehensive experiments on the Cora and CiteSeer datasets, highlighting the significant contributions of our work. The main contributions of this paper are as follows:

(1) We introduce a novel node classification approach leveraging a semantic-structural attention-enhanced graph convolutional network (SSA-GCN) model. This model effectively harnesses the inherent features embedded within the entire graph data, enhancing classification accuracy.

(2) Our proposed model innovatively incorporates the TransE algorithm for knowledge graph embedding to extract rich semantic features. Furthermore, it leverages the node2vec algorithm for graph embedding to capture intricate structural features. These features are then seamlessly fused through a cross-attention mechanism, significantly boosting the model's generalization capabilities.

(3) Our method has demonstrated performance improvement on two benchmark datasets, even under privacy-preserving settings. 

\section{Related Work}
\subsection{Node Classification}
Rong et al. \cite{DBLP:conf/iclr/RongHXH20} introduced the DropEdge method, an innovative technique aimed at mitigating the overfitting issue. This method effectively addresses the problem by randomly removing edges from the input graph data, thereby enhancing the model's generalization capabilities. Hu et al. \cite{DBLP:conf/ijcai/Hu0WWT19} introduce a hierarchical graph convolutional network that significantly expands the receptive field of nodes, enabling it to capture a broader range of global information. This approach enhances the model's ability to capture complex patterns and relationships within the graph data, leading to improved performance in various tasks. Oono et al. \cite{DBLP:conf/iclr/OonoS20} employed spectral graph and random graph theory to delve deeply into the intricacies of graph neural networks. Their analysis not only shed light on the subpar performance of deep nonlinear graph neural networks but also provided valuable insights for the design of improved network architectures. This comprehensive study offers a fresh perspective on enhancing the performance of graph neural networks. Wang et al. \cite{wang2022graphfl} introduced GraphFL, a cutting-edge federated learning framework designed specifically for graph node classification. This framework addresses the challenges associated with non-independent and non-identical distributions in graph data, paving the way for more robust and accurate classification tasks in distributed settings. Zhou et al. \cite{DBLP:conf/cikm/0002CZTZG19} innovatively proposed the application of meta-learning to graph data, offering a solution to the challenging problem of few-sample node classification. This approach leverages the power of meta-learning to enhance the model's generalization capabilities, enabling accurate classification even with limited samples.

Hu et al. \cite{DBLP:journals/corr/abs-2106-04051} assert that both spatial-based and spectral-based graph neural networks heavily rely on the adjacency matrix message-passing mechanism. They propose the utilization of Graph-MLP and neighborhood contrastive loss as alternatives to the traditional message-passing mechanism, potentially revolutionizing the field of graph neural networks. Wu et al. \cite{DBLP:journals/kbs/WuSHYXJ21} introduced a novel approach to modeling the intricate label relationships between node pairs. By leveraging additional supervisory information, they effectively enhanced node classification tasks through multi-task training, offering a powerful framework for improved performance in graph-based learning scenarios. Xiao et al. \cite{DBLP:journals/mva/XiaoWDG22} conducted a comprehensive review of the research on graph neural networks for node classification tasks, offering a deep understanding of the current state-of-the-art in this domain. Chen et al. \cite{DBLP:journals/kbs/ChenHXDHHL22} introduced a novel approach to enhancing node classification and recommendation tasks by meticulously evaluating the quality of neighborhood nodes. This method offers a refined and targeted way to improve the accuracy and effectiveness of graph-based learning tasks. Lin et al. \cite{DBLP:journals/pr/LinZWYWCW23} conducted a thorough investigation into the adversarial attack of graph neural networks, focusing specifically on the robustness of these networks in the context of node classification tasks. Their findings offer valuable insights into enhancing the resilience of graph neural networks against potential attacks. Zhong et al. \cite{DBLP:conf/acl/ZhongCSGW20} introduced TPC-GCN, a novel approach that harnesses the power of GCN to detect post disputes. Their method effectively models topic-post-comment graph data and incorporates pre-trained language models to represent posts, offering a comprehensive and accurate solution to this challenging task. Different from the above models, the public dataset employed in this paper comprises discrete bag-of-words features. The study primarily explores the influence of combining graph embedding with knowledge graph embedding representations on GCN in general scenarios, providing valuable insights into enhancing the performance of graph-based learning frameworks.

\subsection{Graph Neural Network}
Kipf et al. \cite{DBLP:conf/iclr/KipfW17} revolutionized graph data processing by introducing the concept of convolutional neural networks, traditionally used in image processing, to the realm of graph structures. Their proposed GCN framework encodes the local graph structure and node features within the spectral domain, enabling the derivation of embedded node representations that are subsequently utilized for node classification tasks. This approach offers a powerful and effective way to capture the intricate relationships within graph data. The Graph Attention Network (GAT) \cite{velickovic2017graph} employs a self-attention mechanism to differentially weight neighbor nodes in the spatial domain. This mechanism effectively integrates information from these neighboring nodes, resulting in a rich and informative node representation. Graph Autoencoder (GAE) \cite{DBLP:journals/corr/KipfW16a} utilizes a graph convolutional model to encode the nodes within a graph. Subsequently, a decoder is employed to reconstruct the neighborhood of each node. The loss metric is derived from the similarity between the original and reconstructed neighborhoods. When trained in a variational manner, it is referred to as a variational graph autoencoder, which incorporates latent variables to learn graph embeddings. This model effectively leverages latent variables, enabling it to learn interpretable latent representations for undirected graphs.

Graph Spatial-temporal Networks tackle the challenge of dynamic graph structures, where nodes and edges evolve over time, providing a powerful framework for capturing temporal patterns and spatial dependencies. Manessi et al. \cite{manessi2020dynamic} introduced an innovative approach that seamlessly integrates LSTM and GCN to effectively capture both the temporal and spatial structure of graph data, enabling a deeper understanding and analysis of complex network dynamics. Cen et al. developed CogDL \cite{DBLP:journals/corr/abs-2103-00959}, a comprehensive framework designed to integrate various graph neural network models. This platform not only facilitates distributed training but also enables mixed precision training, significantly enhancing the efficiency and scalability of graph-based learning tasks. Zhou et al. \cite{DBLP:journals/aiopen/ZhouCHZYLWLS20} conducted a comprehensive review, highlighting the key research directions, noteworthy progress, and promising future avenues in the field of graph neural networks.

\section{Approach}
\begin{figure}[!h]
\centering
\includegraphics[width=0.99\linewidth]{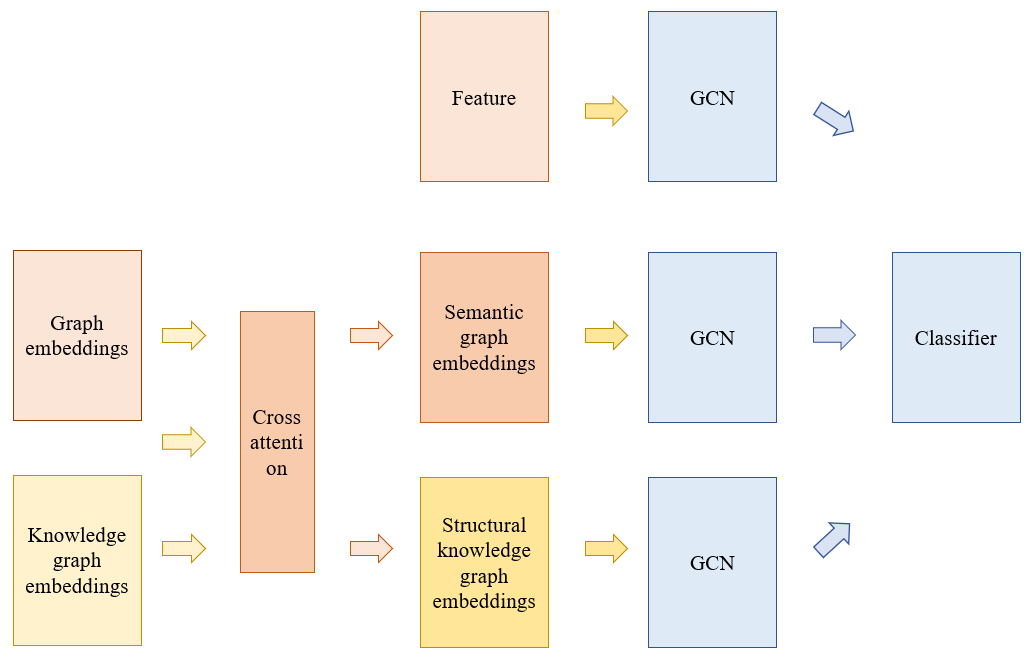}
\caption{Schematic diagram of the model architecture of SSA-GCN}
\label{arch.fig}
\end{figure}

Given a graph dataset G = (V, E), where V denotes the collection of nodes and E represents the set of edges connecting them, the objective of the node classification task is to assign predefined categories Y to each node in V. This assignment is based on both the node features X and the graph structure encoded by E. The innovation of this paper lies in introducing a semantic-structural attention-enhanced graph convolutional network. This approach aims to harness the inherent properties of graph data to refine the parameter optimization process of graph convolutional network models, ultimately enhancing the accuracy of node classification tasks. In the subsequent sections, we delve into the intricacies of the proposed method, highlighting its key components: semantic embedding grounded in knowledge graph embedding techniques, structural embedding leveraging graph embedding algorithms, a cross-attention mechanism, and the overall architecture of the SSA-GCN, as shown in Figure \ref{arch.fig}.

\subsection{Learning Semantic Features}
\label{semantic.sec}


Knowledge graph embedding algorithms are primarily utilized to derive semantic embeddings. Given a set of triples $S=\{(h,r,t)|h,t\in E, r\in R\}$, where $E$ represents the entity set and $R$ represents the relationship type set, the objective of knowledge map embedding is to discover a mapping $f:E,R\xrightarrow{}\mathbb{R}^d$, which encodes the entities and relations in the knowledge graph as low-dimensional dense vectors. This representation enables the knowledge graph to facilitate the use of downstream tasks. Notably, the TransE[4] algorithm seeks to optimize the vector representation such that $h+r\approx t$, while the interval-based sorting loss (triplet ranking loss) can be employed for learning, as depicted below. 

\begin{align}
L=\sum_{(h,r,t)\in S}\sum_{(h',r,t')\in S'}[\gamma + d(h+r, t) - d(h'+r, t')]_{+}
\end{align}
where $\gamma$ is an interval parameter, and $[\cdot]_+$ means that only the positive part is taken. $S'$ represents the set of negative samples, which includes the following elements.
\begin{align}
S'=\{(h',r,t)|h',t \in E\}\cup\{(h,r,t')|h,t' \in E\}
\end{align}
where $h'$ and $t'$ represent the head entity and tail entity obtained through random sampling, where the two entities are not sampled simultaneously.


The TransE algorithm's learning process involves modeling relation types, but many graph datasets lack explicit relation types and only provide adjacency information between nodes. In such cases, this paper assumes the default relation type. For graph datasets that contain multiple relation types, a direct correspondence can be established between them and the knowledge graph's relation types. Moreover, graph data often feature one-to-many or many-to-many relations, which can be addressed using methods such as TransR \cite{DBLP:conf/aaai/LinLSLZ15} and TransD \cite{DBLP:conf/acl/JiHXL015}. These methods have the potential to be explored in future research.


TransE employs an unsupervised learning algorithm to acquire node representations, which captures the semantic relationships between nodes. This is in contrast to traditional GCNs, which rely on supervised learning tasks to passively extract neighborhood information from graph data. By combining knowledge graph embedding representations, the generalization capability of node representation is significantly enhanced. 

\subsection{Learning Structural Features}
\label{structural.sec}
The graph embedding algorithm is primarily utilized to acquire graph structure embeddings, with the aim of discovering a mapping $f: V\xrightarrow{}\mathbb{R}^d$ that transforms the nodes in the graph into compact, low-dimensional dense embeddings. This mapping allows similar nodes in the graph to have similar distances in the low-dimensional space. In this article, we employ the node2vec algorithm, which applies the principles of Skip-gram \cite{DBLP:journals/corr/abs-1301-3781,DBLP:conf/kdd/PerozziAS14} to graph data. The optimization goal of this algorithm is to maximize the probability of observing the neighborhood of the network, given the embedding of the central node u, under the conditional independence assumption and feature space symmetry assumption. The optimized objective function is presented below.

\begin{align}
\max_{f}\sum_{u\in V}[-\log Z_u+\sum_{n_i \in N_s(u)}f(n_i)\cdot f(u)]
\end{align}
where $f(\cdot)$ is the mapping function from node to feature representation. $u$ is a node, $V$ is a node set on the graph, and $N_s(u)$ is a node set obtained by sampling the adjacent node set of u. The calculation method of $Z_u$ is shown below.
\begin{align}
Z_u=\sum_{v\in V} e^{f(u)\cdot f(v)}
\end{align}
Where $u$ represents the central node and v represents any node. The random sampling process is a method of designing a biased random walk on the basis of breadth-first search and depth-first search, as shown below.
\begin{align}
p(c_i = x|c_{i-1}=v)=\begin{cases}\frac{p_{vx}}{C} & (u,v) \in E \\ 0 & \text{otherwise} \end{cases}
\end{align}
$p_{vx}$ is the unnormalized transition probability between nodes v and x, and C is a normalized constant. The $p_{vx}$ transition probability can be designed according to the distance between the previous node and the next node.

Graph embedding representations capture the structural correlation information of graph data through unsupervised learning algorithms. GCNs fail to extract the structural correlations of random walks during supervised learning. Finally, the generalization of node features is enhanced by graph embedding representation.

\subsection{Cross-Attention Mechanism}
\label{attention.sec}
In traditional approaches, structural embedding and semantic embedding often operate independently, lacking deep interactions. However, this paper introduces an attention-enhanced feature fusion technique that leverages a cross-attention mechanism to foster complementarity between semantic and structural embeddings. This fusion method ensures that both types of embeddings contribute synergistically to enhance overall performance. The specific calculation formula for this process is outlined below.
\begin{align}
V^{(g)} = softmax(W^{Q(g)}H^{(kg)} (W^{K(g)}H^{(kg)})^T)H^{(g)}
\end{align}
where $W^{Q(g)}, W^{K(g)}$ are parameter matrices. $H^{(kg)}$ is a collection of knowledge graph embeddings. $H^{(g)}$ is the set of graph embeddings. $V^{(g)}$ represents a set of graph embeddings that have been semantically enhanced.
\begin{align}
V^{(kg)} = softmax(W^{Q(kg)}H^{(g)} (W^{K(kg)}H^{(g)})^T)H^{(kg)}
\end{align}
where $W^{Q(kg)}, W^{K(kg)}$ are parameter matrices. $V^{(kg)}$ denotes a set of knowledge graph embeddings that have been enhanced with structural information. The above embedding representation can be further expanded and enriched through the employment of a multi-head attention mechanism.

\subsection{Semantic-Structural Attention-Enhanced Graph Convolutional Network}
This section delves into the theoretical underpinnings of graph convolutional networks and presents the enhanced approach.
\subsubsection{Graph Convolutional Network}
In spectral-based graph neural network research methods, graph data is treated as undirected. The graph convolutional network model, a specialized form of graph neural network, utilizes the regularized graph Laplacian matrix, which possesses favorable mathematical properties. This matrix serves as a robust mathematical representation of graph data, as outlined below.
\begin{align}
L=I_n - D^{-\frac{1}{2}}A D^{-\frac{1}{2}}
\end{align}
where A is the adjacency matrix, and the diagonal matrix D is the degree matrix. The calculation process is shown below.
\begin{align}
D_{ii} = \sum_{j}A_{i,j}
\label{dii}
\end{align}
The graph Laplacian matrix, exhibiting robust characteristics, is a real symmetric positive semi-definite matrix. Leveraging its mathematical properties, it can be spectrally decomposed, as detailed below.
\begin{align}
L=U\Lambda U^T
\end{align}
where U is a matrix composed of eigenvectors, and the eigenvectors constitute a set of orthogonal basis, and $\Lambda$ is the eigenvalue matrix of L. The calculation process of graph convolution is shown below.
\begin{align}
H^{(l+1)} = \sigma(\widetilde{D}^{-\frac{1}{2}} \widetilde{A} \widetilde{D}^{-\frac{1}{2}} \widetilde{A} H^{(l)}W^{(l)})
\end{align}
Where $\tilde{A}=A+I$. $I$ is the identity matrix. $\widetilde{D}$ is the degree matrix of $\widetilde{A}$, the calculation process is the same as formula \eqref{dii}. $H^{(l)}$ is the hidden state of each layer, $H^{(0)}$ is the original input feature X, $\sigma(\cdot)$ is a nonlinear activation function.

\subsubsection{Semantic-Structural Attention Augmentation Technique}
Existing graph convolutional networks predominantly rely on tailored model architectures to learn task-specific features, often neglecting the targeted extraction of inherent structural and semantic features present in graph data. In this paper, we introduce a semantic-structural graph convolutional network designed to enhance the node classification capabilities of GCN by effectively extracting both semantic embedding and structural embedding features from graph data, as shown below.
\begin{align}
h_i=GCN(x_i, G) \oplus GCN(v_i^{(g)}, G) \oplus GCN(v_i^{(kg)}, G)
\end{align}
where $\oplus$ represents the operation of matrix concatenation. The structural embedding representation is derived from the graph embedding algorithm, whereas the semantic embedding representation stems from the knowledge graph embedding algorithm. The knowledge graph embedding algorithm specializes in capturing the semantic relationships between entities, as detailed in Subsection \ref{semantic.sec}. Conversely, the graph embedding algorithm leverages random walks to learn the structural information encoded within the graph, as explained in Subsection \ref{structural.sec}. By integrating these semantic and structural embeddings into the GCN framework, we can significantly enhance the generalization capabilities of the model.

The ultimate hidden state representation is subsequently projected onto the probability distribution of the various categories via a linear layer, as illustrated below.
\begin{align}
p(y_i)=softmax(W^Th_i)
\end{align}
where $y_i$ represents the probability of each category corresponding to the i-th node. $W$ is a parameter matrix that maps hidden state representations to class probabilities. The objective function for model optimization is the negative log-likelihood loss, as shown below.
\begin{align}
\min_\Theta{-\sum_{i=1}^{|V|}{\log\ p(y_i|h_i;\Theta)}}
\end{align}
Where $\Theta$ represents the model parameters, and $|V|$ represents the number of nodes.

\section{Experiments}
\subsection{Datasets}
The Cora \cite{DBLP:journals/ir/McCallumNRS00} dataset consists of 2708 scientific literature nodes, which are divided into 7 categories. The citation relationship network between literature nodes contains 5429 edges. The feature of each node consists of a 1433-dimensional 0/1 bag-of-words vector, where each position represents the absence/presence of the corresponding word in the bag-of-words. To ensure a more equitable evaluation, we employ random sampling to allocate the training, evaluation, and test sets in a proportion of 8:1:1.

The CiteSeer \cite{DBLP:conf/dl/GilesBL98} data set was proposed by Giles et al., which contains 3312 scientific literature nodes, which are divided into 6 categories. The reference network between nodes contains 4732 edges. The features of a node consist of 3703-dimensional 0/1 bag-of-words vectors, where each position represents the absence/presence of the corresponding word in the bag-of-words. To ensure a more equitable evaluation, we employ random sampling to allocate the training, evaluation, and test sets in a proportion of 8:1:1.

\subsection{Hyper-parameter configuration}
The hidden state of the graph convolutional network is set to 32 dimensions, the learning rate is 0.001, the dropout is 0.2, and the optimization algorithm is Adam \cite{DBLP:journals/corr/KingmaB14}. The vector dimension of the training graph embedding is 128, p=0.25, q=0.25, the learning rate is 0.025, the window size is 10, the random walk length is 80, and the optimization algorithm adopts SGD\cite{DBLP:journals/corr/Ruder16}. The hyperparameters for training knowledge graph embedding are set as batch size 2,000, learning rate 1.0, epoch 2,000, vector dimension 200, and optimization algorithm using Adagrad \cite{lydia2019adagrad}. We randomly executed the program 10 times and calculated the average outcome as the final result. The running environment of the experiment is Intel(R) Xeon(R) Platinum 8163 CPU @ 2.50GHz (Mem: 330G) \& 8 Tesla V100s and an Intel(R) Xeon(R) CPU E5-2680 v4 @ 2.40GHz (Mem: 256) \& 8 RTX 2080Tis.
\subsection{Evaluation}
In this paper, we employ classification accuracy as the primary evaluation metric.

\subsection{Node classification experiment results}
GAT, proposed by Velickovic et al. \cite{velickovic2017graph}, attends to the feature vectors of neighboring nodes and subsequently aggregates these features to capture rich contextual information. On the other hand, SAGE, introduced by Hamilton et al. \cite{hamilton2017inductive}, innovatively divides the convolution process into sampling and aggregation steps, enabling efficient and inductive learning on large-scale graphs. Furthermore, SGC, developed by Wu et al. \cite{wu2019simplifying}, simplifies multi-layer neural networks by collapsing them into a straightforward linear transformation, achieving this by eliminating non-linear activations between GCN layers, thus enhancing efficiency and interpretability.

Table \ref{5nlpres} presents the accuracy achieved on the Cora and CiteSeer datasets, highlighting the best results with bold font. Notably, SSA-GCN outperformed the original GCN by a significant margin of +3.3\% on the Cora test set, while SSA-GAT also exhibited an impressive improvement of +2.6\% compared to the baseline GAT.

On the CiteSeer dataset, SSA-GCN demonstrated a noteworthy improvement of +1.8\% over the original GCN on the test set, while SSA-GAT achieved a slight but positive gain of +0.2\% compared to the baseline GAT. The method introduced in this paper has consistently delivered enhanced performance across both datasets.

\begin{table}[]
\centering
\caption{Accuracy of Various Methods on Cora and CiteSeer Datasets}
\begin{tabular}{c|cc|cc} \toprule
                 & \multicolumn{2}{c|}{Cora} & \multicolumn{2}{c}{CiteSeer}     \\
Models          & Dev        & Test             & Dev    & Test        \\ \midrule
GCN   & 84.8 &  82.8  & 75.8 &   74.5   \\
GAT   & 86.4 &  83.4  & 75.6 &   75.8   \\
SAGE   & 83.4 & 82.8   & 77.1 &  75.1    \\
SGC   & 83.9 & 82.6   & 74.7 &  73.0    \\
\midrule
\textbf{SSA-GCN}  & 85.5 & \textbf{86.1}    & 77.7  & \textbf{76.3}\\ 
\textbf{SSA-GAT}  & \textbf{88.7} & 86.0 & \textbf{78.6}  & 76.0 \\ 
\bottomrule       
\end{tabular}

\label{5nlpres}
\end{table}

\subsection{Analysis of Divergence in Semantic-Structural Embeddings}
\begin{figure}[!h]
\centering
\includegraphics[width=3in]{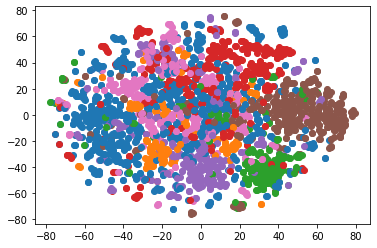}
\caption{Visualization of TransE embeddings}
\label{transe.fig}
\end{figure}

This subsection primarily delves into the characteristics of node embedding distribution, aiming to discern any disparities in intra-class and inter-class divergences. Furthermore, it explores the interpretability of these embeddings. Leveraging the Cora dataset, this article acquires embeddings and employs t-SNE to dimensionally reduce and visualize them. Nodes sharing the same color signify belonging to the same category. As Figure \ref{transe.fig} illustrates, the node representations learned through TransE form distinct clusters exhibiting significant inter-class divergence yet minimal intra-class divergence.

\begin{figure}[!h]
\centering
\includegraphics[width=3in]{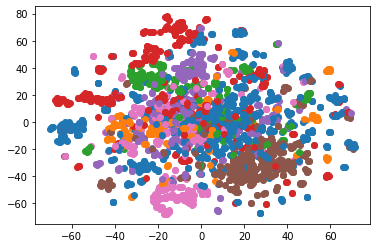}
\caption{Visualization of node2vec embeddings}
\label{ne.fig}
\end{figure}

The embedding representation learned using node2vec exhibits pronounced intra-class aggregation characteristics, evident in Figure \ref{ne.fig}. Experimental results reveal that the unsupervised learning of node embeddings through TransE and node2vec algorithms is effective in capturing categorical features. This attribute is highly advantageous for node classification in privacy-sensitive scenarios.

\subsection{Ablation Study}
We conducted an ablation study on the Cora dataset, and the results are summarized in Table \ref{ablation}. Upon removing the attention module, the experimental performance decreased by -0.3\%. When the knowledge graph embedding module was also eliminated, the results further declined by -0.6\%. Finally, upon further removing the graph embedding module, the experimental results decreased significantly by -3.3\%. These findings indicate that, among the modules introduced in this paper, graph embedding contributes the most significantly, while the other modules also make notable contributions.
\begin{table}[!htbp]
\caption{Ablation study results on the Cora dataset}
\small
\centering
    \begin{tabular}{l|cc}
    \toprule
    Settings          & Dev  & Test    \\ 
    \midrule
    SSA-GCN & 85.5  & 86.1  \\
    -- Attention & 86.9 & 85.8 \\
    -- Attention, KGE & 86.3 & 85.5 \\
    -- Attention, KGE, GE & 84.8 & 82.8 \\
    \bottomrule
    \end{tabular}

\label{ablation}
\end{table}

\subsection{Classification Results under Privacy Settings}
We conducted experiments under privacy settings to assess the impact of node classification in scenarios where node data features are unavailable. In practical applications, machine learning systems often encounter privacy protection and authorization challenges with sensitive data, thus hampering node classification tasks. The absence of raw node features can render graph neural networks ineffective. This paper aims to address the challenge of node classification in the absence of original node features. To simulate a scenario with limited data availability, we masked the original features of nodes and solely relied on the learned embedding representation to infer node types.

Table \ref{privacy} presents the experimental results obtained on the Cora dataset, with the best results highlighted in bold. privacy-GCN+KGE refers to the approach where only knowledge graph embeddings are used as features, while privacy-GCN+GE utilizes solely graph embeddings. privacy-GCN+KGE, GE incorporates both knowledge graph embeddings and graph embeddings as features. privacy-SSA-GCN and privacy-SSA-GAT represent the SSA-GCN and SSA-GAT methods introduced in this article, respectively, where the original node features are excluded. The experimental results demonstrate that our proposed method can achieve a maximum accuracy of 79.6\%, which is only a 6.1\% decrease compared to utilizing the original node features. This indicates that node information can be effectively inferred through semantic and structural embeddings. Therefore, the method proposed in this article is a viable option for node classification tasks in privacy-sensitive scenarios.

\begin{table}[]
\centering
\caption{Accuracy of Cora Dataset under Privacy Settings}
\begin{tabular}{c|cc}   \toprule
Models          & Dev        & Test                   \\ \midrule
privacy-GCN + KGE  & 67.1  &   65.5    \\ 
privacy-GCN + GE  & 78.1  &  76.4    \\ 
privacy-GCN + KGE + GE &  \textbf{80.9} &  77.4   \\ \midrule
\textbf{privacy-SSA-GCN}  &  79.1 &  77.6    \\ 
\textbf{privacy-SSA-GAT}  &  79.1 & \textbf{79.6}    \\ 
\bottomrule

\end{tabular}
\label{privacy}
\end{table}

\subsection{Conclusion}
This paper introduces a semantic-structural attention-enhanced graph convolutional network, specifically designed to extract inherent semantic-structural features from comprehensive graph data, thereby enhancing the model's generalization capabilities. For semantic embedding, we employ the TransE knowledge graph embedding algorithm to capture semantic information from a knowledge graph perspective through unsupervised feature extraction. For structural embedding, we utilize the node2vec graph embedding algorithm to extract structural information from a complex network perspective, also through unsupervised feature extraction. In this paper, we integrate both semantic and structural embeddings into the graph convolutional network, leveraging a cross-attention mechanism to merge these features, ultimately enhancing node classification performance.

The innovation of this paper lies in the utilization of a cross-attention mechanism to seamlessly integrate semantic and structural embeddings within graph convolutional networks, thereby enhancing the performance of node classification tasks. We envision applying our model to the analysis of large-scale graph data in the future.

\bibliographystyle{IEEEtran}
\bibliography{reference}

\end{document}